\documentclass[sigconf]{acmart}

\usepackage{booktabs} 
\usepackage{color}

\newcommand*{\affmark}[1][*]

\setcopyright{none}

\newcommand{\etal}{\mbox{\emph{et al.}}}

\begin{document}
\title{Self-Supervised Visual Representations for Cross-Modal Retrieval}

\author{Yash Patel}
\affiliation{%
  \institution{The Robotics Institute, Carnegie Mellon University, PA, USA}
}
\email{yashp@andrew.cmu.edu}

\author{Lluis Gomez}
\affiliation{%
  \institution{Computer Vision Center, Universitat Autonoma de Barcelona, Spain}
}
\email{gomez@cvc.uab.es}



\author{Mar\c{c}al Rusi\~{n}ol}
\affiliation{%
  \institution{Computer Vision Center, Universitat Autonoma de Barcelona, Spain}
}
\email{marcal@cvc.uab.es}

\author{Dimosthenis Karatzas}
\affiliation{%
  \institution{Computer Vision Center, Universitat Autonoma de Barcelona, Spain}
}
\email{dimos@cvc.uab.es}

\author{C.V. Jawahar}
\affiliation{%
  \institution{CVIT, KCIS, IIIT Hyderabad, India}
}
\email{jawahar@iiit.ac.in}

\renewcommand{\shortauthors}{Y. Patel ~\etal}

\begin{abstract}

Cross-modal retrieval methods have been significantly improved in last years with the use of deep neural networks and large-scale annotated datasets such as ImageNet and Places. However, collecting and annotating such datasets requires a tremendous amount of human effort and, besides, their annotations are usually limited to discrete sets of popular visual classes that may not be representative of the richer semantics found on large-scale cross-modal retrieval datasets. In this paper, we present a self-supervised cross-modal retrieval framework that leverages as training data the correlations between images and text on the entire set of Wikipedia articles. Our method consists in training a CNN to predict: (1) the semantic context of the article in which an image is more probable to appear as an illustration (global context), and (2) the semantic context of its caption (local context). Our experiments demonstrate that the proposed method is not only capable of learning discriminative visual representations for solving vision tasks like image classification and object detection, but that the learned representations are better for cross-modal retrieval when compared to supervised pre-training of the network on the ImageNet dataset.

\end{abstract}

%
%


\keywords{Self-Supervised Learning, Visual Representations, Cross-Modal Retrieval}

\maketitle

\section{Introduction}
\label{sec:introduction}

\begin{figure*}
\includegraphics[width=\textwidth]{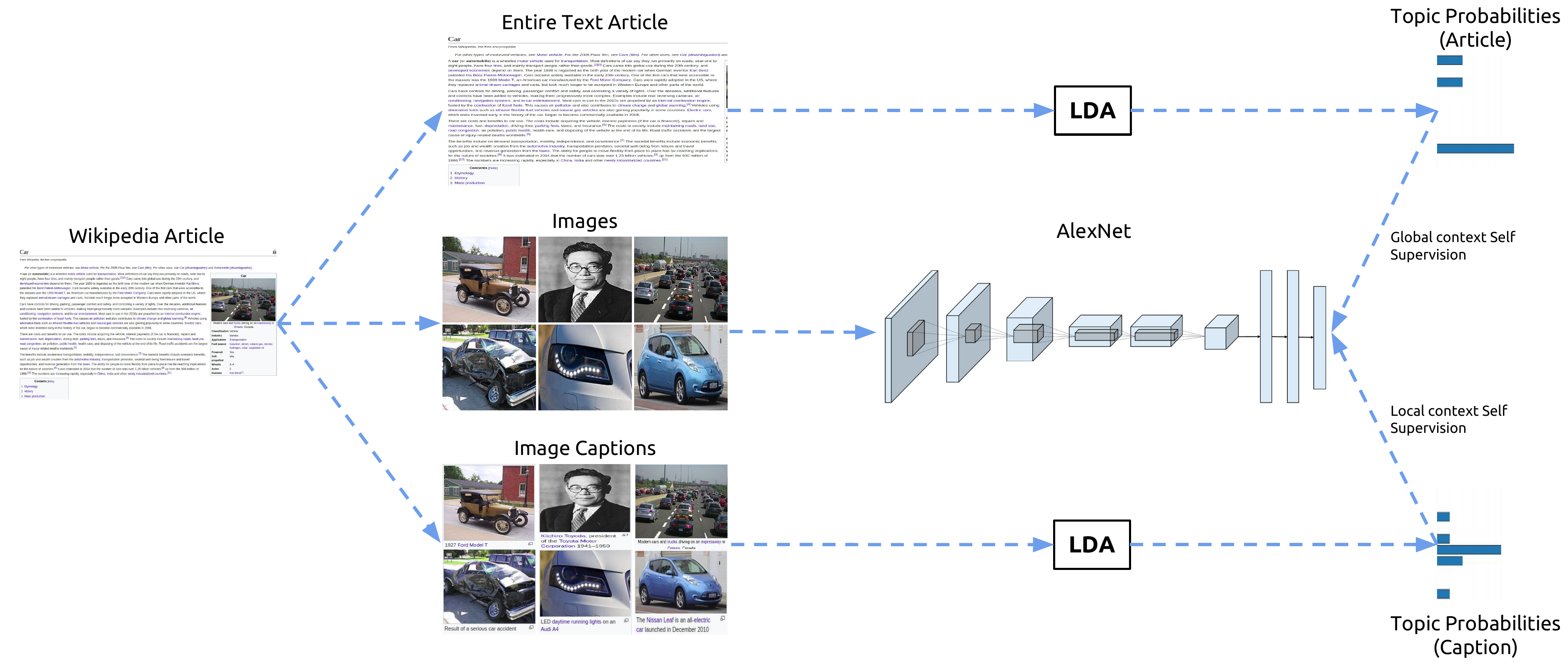}
\caption{Method overview: Wikipedia articles contain textual description of a subject, these articles are also accompanied with illustrative images supporting the text. These images are often accompanied by captions. A Latent Dirichlet Allocation (LDA) \cite{blei2003latent} topic modeling framework generates a global contextual representation of the textual information from entire text article. The same LDA model generates a local contextual representation from the per-image caption. These two text representations are jointly used to supervised the training of deep CNN.}
\label{fig:overview}
\end{figure*}

The emergence of large-scale annotated datasets such as ImageNet~\cite{deng2009imagenet}, Places~\cite{zhou2014learning} and MS-COCO~\cite{lin2014microsoft} has undoubtedly been one of the key ingredients for the tremendous impact of deep learning on almost every computer vision task. However, there is a major issue with the supervised learning setup in large scale datasets; collecting and manually annotating those datasets requires a great amount of human effort. On the other hand, the fact that the annotations on such datasets are usually limited to discrete sets of popular visual classes, may not necessarily be an optimal training setup for cross-modal retrieval datasets that usually cover a set of broader and richer semantic concepts.

As an alternative to the fully supervised setup, self-supervised learning methods aim at learning discriminative visual features by designing auxiliary tasks for which the target labels are free to obtain. These labels provide supervision for the training of computer vision models the same way as in supervised learning, but the supervisory signal can be directly obtained from the training data, either from the images themselves~\cite{doersch2015unsupervised,pathak2016context} (uni-modal training) or from a complementary modality that is found naturally correlated with them ~\cite{agrawal2015learning,owens2016ambient,gomez2017self} (multi-modal training). Unlike supervised learning, where visual features are learnt from human generated labels, in self-supervised learning labels are automatically obtained from the training data.

In this paper we present a self-supervised cross-modal retrieval framework that leverages as supervisory signal the correlations found between images and text on a large collection of illustrated articles in order to learn discriminative visual features that could potentially transfer well to any general computer vision task, such as image classification or object detection. We hypothesise that the learned representations by using such approach can be used more naturally in a cross-modal retrieval framework than the representations learned from annotated datasets for image classification.

Our intuition follows from the observation that illustrated encyclopedic articles, like Wikipedia's ones, are well organized and contain a detailed textual description of their subject while certain aspects of the subject are illustrated by images.
Those images complement the text and at the same time provide context to our imagination. Furthermore, the captions associated with these images specifically describe their contents. These observations, and the large-scale availability of such articles, lead us to treat representation learning for cross-modal retrieval as a self-supervised visual representation learning task. We demonstrate that rich visual representations can be learned by training a network to predict the global (article-level) and local (caption-level) semantic contexts in which an image appears, and at the same time, the learned representations can be used to perform cross-modal retrieval with promising results.

Gomez and Patel~\etal~\cite{gomez2017self,patel2018texttopicnet} have proposed in the past self-supervised representation learning using Wikipedia articles. Their method consists in learning a Latent Dirichlet Allocation (LDA) model from the entire corpus of text articles, and then training a CNN to predict the semantic context of images by using as training labels the semantic level representations (the probability distribution over semantic topics) of the articles in which they appear, as provided by the LDA model. An assumption made in their method is that all the images within a given text article have the same target semantic representation, which is obtained from the LDA model. However, images within a Wikipedia article can be drastically different in terms of appearance and semantic content. To overcome this, we create a new \emph{Wikipedia dataset with captions} which is similar to the one used in the TextTopicNet~\cite{gomez2017self,patel2018texttopicnet} method, but also containing the image captions from Wikipedia. Thus, as illustrated in Figure~\ref{fig:overview}, the training data in our method comes in a triplet form (image, text article, image caption). 

Our intuition is that adding another target representation based on image captions could provide more image specific training self-supervision. Furthermore, we experimentally show that our training procedure leads to significantly better results for both cross-modal retrieval and image classification. 

Following are the major contributions made in this paper:
\begin{itemize}
\item We propose a multi-task learning framework to train a CNN that predicts text representations obtained from text articles (global context) and per-image captions (local context).
\item We experimentally demonstrate that the self-supervisedly learned visual features are generic enough for other computer vision tasks and outperform other self-supervised and naturally supervised approaches on standard benchmarks.
\item Without using any form of semantic information, our method outperforms both unsupervised and supervised approaches on cross-modal retrieval (image-to-text and text-to-image) benchmarks on Wikipedia \cite{rasiwasia2010new} and Pascal sentences datasets \cite{farhadi2010every}.
\item The Wikipedia image-article dataset \cite{patel2018texttopicnet} consist of only images and text articles, and as an auxiliary contribution, we release a large scale dataset obtained from English Wikipedia consisting of images, per-image captions and co-occurring text articles.
\end{itemize}

\begin{figure*}
\includegraphics[width=\textwidth]{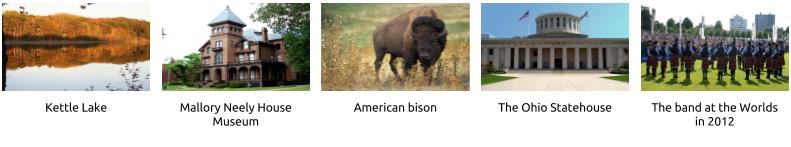}
\caption{Samples from Wikipedia dataset with captions. In the method, the shown captions provide local image specific information, whereas the entire text article provides global subject information.}
\label{fig:dataset_samples}
\end{figure*}

\begin{figure*}
    \centering
    \includegraphics[width=0.8\textwidth]{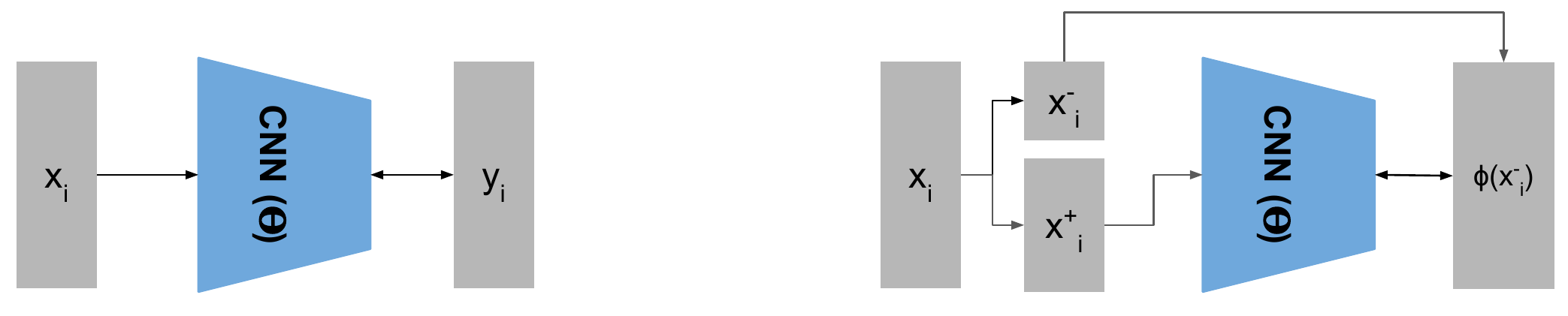}
    \caption{Supervised (left) vs. Self-Supervised Training (right). In supervised training the ground-truth labels $y
    _{i}$ are collected by human annotation. Whereas in self-supervised training, a transformation on a part of input data is used as the target label for training.}
    \label{fig:supervised_self_supervised}
\end{figure*}

The rest of the paper is structured as follows. In Section \ref{sec:related_work}, previous work is reviewed. In Section \ref{sec:dataset}, details of training dataset are elaborated. In Section \ref{sec:method} the proposed method is described and in Section \ref{sec:experiments} evaluated. The paper is concluded in Section \ref{sec:conclusion}.

\section{Related Work}
\label{sec:related_work}

\subsection{Self-Supervised Visual Representations}
As an alternative to fully-supervised algorithms, there has recently been a growing interest in self-supervised or naturally-supervised approaches that make use of non-visual signals, intrinsically correlated to the image, as a form to supervise visual feature learning. The objective of those methods is to learn visual representations (without human annotations) that are generic to work well across a wide range of object classes and at the same time are discriminating enough to be useful for classical computer vision tasks such at image classification, object detection, semantic segmentation etc.

\subsubsection{Unsupervised Visual Representation Learning}

Work in unsupervised data-dependent methods for learning visual features has been mainly focused on algorithms that learn filters one layer at a time. A number of unsupervised algorithms have been proposed to that effect, such as sparse-coding, restricted Boltzmann machines (RBMs), auto-encoders~\cite{zhao2015stacked}, and K-means clustering~\cite{coates2010analysis,dundar2015convolutional,krahenbuhl2015data}. However, despite the success of such methods in several unsupervised learning benchmark datasets, a generic unsupervised method that works well with real-world images does not exist. 

Bojanowski \& Joulin \etal~\cite{bojanowski2017unsupervised} present an approach for unsupervised learning of visual features using Noise As Target (NAT) label for training. Their approach is domain agnostic and makes use of fixed set of target labels for training. The primary difference between our work and \cite{bojanowski2017unsupervised} is that, in our work the final network predictions are directly useful for a specific task - cross-modal matching and retrieval.

\subsubsection{Uni-modal Self-Supervised Methods}

In contrast to the purely supervised approaches, uni-modal self-supervised algorithms make use of the structure in the visual data itself for the purpose of representation learning. Agrawal \etal~\cite{agrawal2015learning} make use of egomotion information obtained by odometry sensors mounted on a vehicle. They train a network using contrastive loss formulation \cite{mobahi2009deep} to predict the camera transformations between two image pairs. 

Wang and Gupta \etal~\cite{wang2015unsupervised,wang2017transitive} make use of videos as training data and use relative motion of objects as supervisory signal for training. The relative motion information is obtained by using a standard unsupervised tracking algorithm. A Siamese-triplet network is then trained using a ranking loss function.

Pathak and Efros \etal~\cite{pathak2016context} take inspiration from auto-encoders and proposed a context-encoder. They train a network using a combination of L2 loss and adversarial loss to generate arbitrary image regions conditioned on their surrounding. Doersch \etal~\cite{doersch2015unsupervised} use spatial context such as relative position of patches within an image to make the network learn object and object parts.

Our proposed method is different from all of these methods since it makes use of multi-modal axillary task for training. Further, by training the network to predict local and global contexts in which an image appears as illustration could be directly used for cross-modal retrieval. Our work is more correlated with the multi-modal self-supervised approaches as elaborated in next section.

\subsubsection{Multimodal Self-Supervised Methods} 

Multi-modal self-supervised learning alrogithms attempt to utilize the structure in one modality to provide the training supervision for co-occuring modality.

Owens \etal~\cite{owens2016ambient} make use of sound as a modality to provide supervisory signal. They do so by training a deep CNN to predict a hand-crafted statistical summary of sound associated with a video frame.


Gomez and Patel ~\etal \cite{gomez2017self} make use of Wikipedia documents which consist of text articles and co-occuring images. First, a Latent Drichilet Allocation (LDA) \cite{blei2003latent} topic model is learned on the entire Wikipedia dataset. Second, text articles are represented in the form of topic-probabilities using learned LDA model. Finally, a convolutional neural network is trained on images in Wikipedia, where the target label is the representation of corresponding text article.

Our work is more closely related to \cite{gomez2017self,patel2018texttopicnet,patel2016dynamic}, however, as previously mentioned, their approach makes use of same target representation for all images within a text article. This not only leads to sub-optimal performance but also completely ignores the local context of an image.

\subsection{Cross-Modal Representation Learning}

Two general categories of the representation learning methods for cross-modal retrieval can be: (a) \textit{real-valued}, (b) \textit{binary valued}. The binary methods are more focused on efficiency and aim to map the items from different modalities on a common binary hamming space \cite{zhu2013linear,song2013inter,shen2016fast,xu2017learning}.

Our approach falls in the category of \textit{real-valued} methods. Within this category of methods the training for cross-modal retrieval could be: unsupervised \cite{andrew2013deep,hardoon2004canonical,srivastava2012multimodal,feng2014cross,yan2015deep} or supervised \cite{gong2014multi,wang2016joint,wang2013learning,zhai2014learning}.

Zhang ~\etal \cite{zhao2015stacked} propose a multimodal hashing method, called semantic correlation maximization (SCM), which integrates semantic labels into the hashing learning procedure. This method uses label vectors to get semantic similarity matrix and tries to reconstruct it through the learned hash codes.

Gong ~\etal \cite{gong2014multi} propose a novel three-view CCA (CCA-3V) framework, which explicitly incorporates the dependence of visual features and text on the underlying semantics.

Wang ~\etal \cite{wang2013learning} propose a novel regularization framework for the cross-modal matching problem, called LCFS (Learning Coupled Feature Spaces). It unifies coupled linear regressions, $l_{21}$-norm and trace norm into a generic minimization formulation so that subspace learning and coupled feature selection can be performed simultaneously. Furthermore, they extend this framework to more than two-modality case in \cite{wang2016joint}, where the extension version is called JFSSL (Joint Feature Selection and Subspace Learning). 

Wang ~\etal \cite{wang2017adversarial} propose an adversarial learning approach for cross-modal retrieval. The method is built around the idea of min-max game involving two different processes $players$: a modality classifier distinguishing the items in terms of their modalities, and a feature projector generating modality-invariant and discriminate representations and aiming to confuse the modality classifier.

While most of these supervised or unsupervised approaches attempts to learn a common embedding space for the prupose of cross-modal retrieval, they assume that the visual representations are provided by a pre-trained CNN (either AlexNet \cite{krizhevsky2012imagenet} or VGG-16 \cite{simonyan2014very}) on ImageNet dataset \cite{russakovsky2015imagenet}. The cost (human annotation effort) of this pre-training is not accounted by cross-modal retrieval methods. Further, the underlying assumption of is that ImageNet pre-trained features transfer well for cross-modal retrieval.

The proposed method investigates these two aspects, firstly, we do not make use of ImageNet pre-training and instead use the self-supervised visual representations. Secondly, in the experiments, we train the network just once on our dataset and no form of adaptation is done on test datasets. This demonstrates that our proposed method is capable of learning a general purpose category-agnostic cross-modal retrieval system.


\section{Wikipedia Dataset with Captions}
\label{sec:dataset}

In order to obtain a training dataset for our method, we scrapped the entire English Wikipedia while considering only articles with at least 50 words and illustrated with at-least one image. Similarly to the preprocessing of ImageCLEF dataset we filtered small images (< 256 pixels) and images with formats other than JPG. Furthermore, we only considered the images for which captions are available. With these constraints our dataset consists of $1.8$ million images with captions and the associated text article that they appear with. Figure \ref{fig:dataset_samples} shows samples from the dataset.

\section{Method}
\label{sec:method}

In this section, we first elaborate over the core distinction between supervised and self-supervised trainings. Then we discuss about the Latent Dirichlet Allocation (LDA) \cite{blei2003latent}, which is used for representing text articles and image captions, and thus for generating target representations for training the CNN. Finally, we go over the training of the CNN.

\subsection{Self-Supervised Learning}
\label{sec:self_supervised}

The supervised methods learn rich visual representations from large collections of training data. This data always 
has 
human annotations, $D=\{(x_{1},y_{1}), (x_{2},y_{2}) ... (x_{N},y_{N})\}$, and the deep network is trained to minimize the overall risk term:

\begin{equation}
\small
R=\sum_{i=1}^{N}[loss(f(x_{i},\Theta),y_{i})]
\end{equation}

Where $\Theta$ are the parameters of the deep network.

Unlike supervised approaches, self-supervised methods train without making use of any human annotations. The training data, $D=\{(x_{1}), (x_{2}) ... (x_{N})\}$ can be sub-divided into components and one or more components can be used to provide self-supervision for others, thus, data
is represented as $D=\{(x_{1}^{+}, x_{1}^{-}), (x_{2}^{+}, x_{2}^{-}) ... (x_{N}^{+}, x_{N}^{-})\}$ and the training for one component 
is governed by the other changing the overall risk term to:

\begin{equation}
\small
R=\sum_{i=1}^{N}[loss(f(x_{i}^{+},\Theta),x_{i}^{-})]
\end{equation}

Fig.~\ref{fig:supervised_self_supervised} shows the explicit difference between supervised and self-supervised approaches.

\subsection{Latent Dirichlet Allocation}
\label{sec:text_rep}
LDA \cite{blei2003latent} is a generative statistical model of a text corpus where each document can be viewed as a mixture of various topics, and each topic is characterized by a probability distribution over words. LDA can be represented as a three level hierarchical Bayesian model. Given a text corpus consisting of $M$ documents and a dictionary with $N$ words, Blei \etal define the generative process~\cite{blei2003latent} for a document $d$ as follows:
\begin{itemize}
\item{Choose $\theta \sim Dirichlet(\alpha)$.}
\item{For each of the $N$ words $w_n$ in $d$:}
 \begin{itemize}
 \item{Choose a topic $z_n \sim Multinomial(\theta)$.}
 \item{Choose a word $w_n$ from $P(w_n \mid z_n, \beta)$, a multinomial probability conditioned on the topic $z_n$.}
 \end{itemize}
\end{itemize}
\noindent
where $\theta$ is the mixing proportion and is drawn from a Dirichlet prior with parameter $\alpha$, and both $\alpha$ and $\beta$ are corpus level parameters, sampled once in the process of generating a corpus. Each document is generated according to the topic proportions $z_{1:K}$ and word probabilities over $\beta$. The probability of a document $d$ in a corpus is defined as: 
\small
\begin{equation}
P(d\mid\alpha, \beta) =  \nonumber
\int_{\theta}P(\theta \mid\alpha)\left(\prod_{n=1}^{N}\sum_{z_{K}}^{ } P(z_{K} \mid \theta)P(w_{n}\mid z_{K},\beta)\right)d\theta \nonumber
\end{equation}
\normalsize
Learning LDA~\cite{blei2003latent} on a document corpus provides two set of parameters: word probabilities given topic $P(w\mid z_{1:K})$ and topic probabilities given document $P(z_{1:K} \mid d)$. Therefore each document is represented in terms of topic probabilities $z_{1:K}$ (being $K$ the number of topics) and word probabilities over topics. Any new (unseen) document can be represented in terms of a probability distribution over topics of the learned LDA model by projecting it into the topic space.

\subsection{Network Architecture}
\label{sec:network_architecture}
Throughout our experiments, we make use of AlexNet architecture \cite{krizhevsky2012imagenet}. The choice of AlexNet is justified
because most of the existing self-supervised methods make
use of this same architecture \cite{wang2016unsupervised,agrawal2015learning,owens2016ambient,gomez2017self,patel2018texttopicnet,pathak2016context}. Further, we compare to cross-modal retrieval methods with reported performance using AlexNet \cite{hardoon2004canonical,rasiwasia2010new,sharma2012generalized,gong2014multi,wang2013learning,wang2016joint}. Thus the use of AlexNet architecture is essential for fair comparisons.  

As shown in Figure \ref{fig:overview}, till the \textit{fc7} layer the architecture is same as standard AlexNet \cite{krizhevsky2012imagenet}, which is followed by two fully-connected branches one prediction caption level topic probabilities and other predicting article level topic probabilities.

\begin{table*}
\resizebox{\textwidth}{!}{
\begin{tabular}{l | c | c | c | c | c | c | c | c | c | c | c | c | c | c | c | c | c | c | c | c }
\toprule
Method &aer &bk &brd &bt &btl &bus &car &cat &chr &cow &din &dog &hrs &mbk &prs &pot &shp &sfa &trn &tv \\
\midrule
Ours &\textbf{73} &\textbf{56} &\textbf{49} &\textbf{65} &\textbf{26} &\textbf{50} &\textbf{73} &\textbf{46} &\textbf{48} &\textbf{38} &\textbf{45} &\textbf{42} &73 &\textbf{64} &\textbf{86} &\textbf{34} &\textbf{44} &\textbf{44} &\textbf{74} &\textbf{48} \\
\midrule
TextTopicNet (Wikipedia) \cite{patel2018texttopicnet} & 71 & 52 & 47 & 61 & 26 & 49 & 71 & 46 & 47 & 36 & 44 & 41 & 72 & 62 & 85 & 31 & 40 & 42 & 72 & 44\\
TextTopicNet (ImageCLEF) \cite{gomez2017self} &67 &44 &39 &53 &20 &49 &68 &42 &43 &33 &41 &35 &70 &57 &82 &30 &31 &39 &65 &41 \\
Sound~\cite{owens2016ambient} &69 &45 &38 &56 &16 &47 &65 & 45 &41 &25 &37 &28 &\textbf{74} &61 &85 &26 &39 &32 &69 &38 \\
Texton-CNN &65 &35 &28 &46 &11 &31 &63 &30 &41 &17 &28 &23 &64 &51 &74 &9 &19 &33 &54 &30 \\
K-means &61 &31 &27 &49 &9 &27 &58 &34 &36 &12 &25 &21 &64 &38 &70 &18 &14 &25 &51 &25\\
Motion~\cite{wang2015unsupervised} &67 &35 &41 &54 &11 &35 &62 &35 &39 &21 &30 &26 &70 &53 &78 &22 &32 &37 &61 &34 \\
Patches~\cite{doersch2015unsupervised} &70 &44 &43 &60 &12 &44 &66 &52 &44 &24 &45 &31 &73 &48 &78 &14 &28 &39 &62 &43 \\
Egomotion~\cite{agrawal2015learning} &60 &24 &21 &35 &10 &19 &57 &24 &27 &11 &22 &18 &61 &40 &69 &13 &12 &24 &48 &28 \\
\midrule
ImageNet~\cite{krizhevsky2012imagenet} &79 &\textbf{71} &\textbf{73} &75 &\textbf{25} &60 &80 &\textbf{75} &51 &\textbf{45} &60 &\textbf{70} &\textbf{80} &\textbf{72} &\textbf{91} &42 &\textbf{62} &56 &82 &62 \\
Places~\cite{zhou2014learning} &\textbf{83} &60 &56 &\textbf{80} &23 &\textbf{66} &\textbf{84} &54 &\textbf{57} &40 &\textbf{74} &41 &\textbf{80} &68 &90 &\textbf{50} &45 &\textbf{61} &\textbf{88} &\textbf{63} \\
\bottomrule
\end{tabular}}
\vspace{0.25em}
\caption{PASCAL VOC2007 per-class average precision (AP) scores for the classification task with pool5 features.}
\label{tab:pascal_pool5_AP}
\end{table*}

\subsection{Learning Self-Supervised Representations}
Following up with the formal definition of self-supervised learning as described in Section \ref{sec:self_supervised}. The multimodal document from Wikipedia can be thought as a training sample, $x_{i}$. This multimodal document consists of text article $x_{i}^{A}$, image captions $x_{i}^{C}$ and images $x_{i}^{I}$. 

Let $\Phi(x_{i}^{A})$ and $\Phi(x_{i}^{C})$ be the text topic probability distributions given by LDA \ref{sec:text_rep} for the document text and the image captions accordingly.
The deep CNN is trained to predict the above topic distributions given the corresponding article image, and producing as outputs: $f_{A}(x_{i}^{I},\Theta)$ (for article) and $f_{C}(x_{i}^{I},\Theta)$ (for caption). 

The loss is computed as the cross entropy between the LDA topic distribution and the predicted distribution. The overall risk term on the training data will be:
\begin{equation}
\small
\begin{split}
R = \sum_{i=1}^{i=N}\lbrack\sum_{topic=1}^{topic=K}\Phi(x_{i}^{A})_{topic}log(f_{A}(x_{i}^{I},\Theta)_{topic}) \\ + \sum_{topic=1}^{topic=K}\Phi(x_{i}^{C})_{topic}log(f_{C}(x_{i,topic}^{I},\Theta)_{topic})\rbrack
\end{split}
\label{eq:overall_risk}
\end{equation}

where $N$ is total number of samples in the training data, $K$ is the number of topics in the LDA model \cite{blei2003latent} and $\Theta$ maps to the learnt CNN parameters. Note that $K$ is a hyper-parameter and we fix $K=40$ throughout the experiments.

\subsection{Training Details}
\label{sec:training_details}
Learning to predict the target topic probability distributions we minimize a sigmoid cross-entropy loss as shown in the overall risk term Eq. \ref{eq:overall_risk}. We use a Stochastic Gradient Descent (SGD) optimizer, with base learning rate of  $0.001$, with a step decay after every $200,000$ iterations by a factor of $0.1$, and momentum of $0.9$. The batch size is set to $128$. With these settings the network converges after $500,000$ iterations of training.

\section{Experiments}
\label{sec:experiments}


We will first compare the learnt visual representations with other self-supervised methods on the task on image classification on two standard benchmark datasets (Section \ref{sec:image_classification}). Next, we will compare our method with various cross-modal retrieval methods (Section \ref{sec:cross_modal_exp}).

\subsection{Self-Supervised Features for Image Classification}
\label{sec:image_classification}

\subsubsection{PASCAL VOC}

Self-supervised learned features are tested for image classification on PASCAL VOC 2007 \cite{everingham2010pascal} dataset. In total there are 9,963 images, and 20 semantic classes. The data has been split into 50\% for training/validation and 50\% for testing. The classification here is multi-label, that is, each image can be classified into multiple classes.

We extract features from the top layers of the CNN (fc7, fc6, pool5) for each image of the dataset. Then, for each class we perform a grid search over the parameter space of an one-vs-all Linear SVM classifier~\footnote{Liblinear implementation from \url{http://scikit-learn.org/}} to optimize its validation accuracy. Then, we use the best performing parameters to train again the one-vs-all SVM using both training and validation images.

\begin{table}
\centering
\begin{tabular}{ l | c | c | c | c}
\toprule
Method & max5 & pool5 & fc6 & fc7 \\
\midrule
Ours & - & \textbf{53.8} & \textbf{54.9} & \textbf{56.8}	\\
\midrule
TextTopicNet (Wikipedia) \cite{patel2018texttopicnet} & - & 51.9 & 54.2 & 55.8\\
TextTopicNet (ImageCLEF) \cite{gomez2017self} & - & $47.4$ & $48.1$ & $48.5$ \\
Sound ~\cite{owens2016ambient} & 39.4 & 46.7 & 47.1 & 47.4 \\
Texton-CNN & 28.9 & 37.5 & 35.3 & 32.5 \\
K-means~\cite{krahenbuhl2015data} & 27.5 & 34.8 & 33.9 & 32.1 \\
Tracking~\cite{wang2015unsupervised} & 33.5 & 42.2 & 42.4 & 40.2 \\
Patch pos.~\cite{doersch2015unsupervised} & 26.8 & 46.1 & - & - \\
Egomotion~\cite{agrawal2015learning} & 22.7 & 31.1 & - & - \\
\midrule
ImageNet~\cite{krizhevsky2012imagenet} & \textbf{63.6} & \textbf{65.6} & \textbf{69.6} & \textbf{73.6} \\
Places~\cite{zhou2014learning} & 59.0 & 63.2 & 65.3 & 66.2 \\
\bottomrule
\end{tabular}
\vspace{0.25em}
\caption{PASCAL VOC2007 mAP comparison for image classification with supervised (bottom), and self-supervised (middle) methods. 
}
\label{pascal_SVM_mAP}
\end{table}

In Tables \ref{tab:pascal_pool5_AP} and \ref{pascal_SVM_mAP}, we compare our results on the PASCAL VOC2007 test set with different state-of-the-art self-supervised learning algorithms using features from different top layers and SVM classifiers. 

Our method which leverages global and local contexts for self-supervised training achieves state-of-the-art performance as seen in Table \ref{pascal_SVM_mAP}. This demonstrates that a network that identifies global and local semantic contexts in which an image is more probable to appear gives better visual representations.

In Table \ref{tab:pascal_pool5_AP}, we provide a per-class comparison with various self-supervised and supervised visual representation learning algorithms. It can be clearly seen that our method performs better than other self-supervised methods for most of the classes. In the case of ``bottle'' class our method outperforms fully supervised network.

\subsubsection{SUN 397}

Table~\ref{tab:sun_SVM_mAP} compares our results on the SUN397 \cite{xiao2010sun} test set with state-of-the-art self-supervised learning and supervised algorithms. SUN397 \cite{xiao2010sun} consists of 50 training and 50 test images for each of the 397 scene classes. We follow the same evaluation protocol as \cite{owens2016ambient,agrawal2015learning} and make use 20 images per class for training and remaining 30 for validation. We evaluate our method on three different partitions of training and testing and report the average performance. This scene classification dataset is suitable for the evaluation of self-supervised approaches as it contains less frequently occurring classes and thus is more challenging compared to PASCAL VOC 2007 dataset.

We appreciate that our method outperforms all other modalities of supervision in this experiment. We observe that using features from \textit{fc6} layer gives better performance compared to using features from \textit{fc7} layer. This indicates that \textit{fc6} and \textit{pool5} layers of our network are more robust towards uncommon classes.

\begin{table}
\begin{tabular}{ l | c | c | c | c}
\toprule
Method & max5 & pool5 & fc6 & fc7 \\ 
\midrule
Ours & - & \textbf{30.3} & \textbf{33.5} & \textbf{28.2} \\
\midrule
TextTopicNet (Wikipedia) \cite{patel2018texttopicnet} & - &  28.8 &   32.2 & 27.7 \\
Sound ~\cite{owens2016ambient} & 17.1 &  22.5 & 21.3 & 21.4  \\
Texton-CNN & 10.7 & 15.2 & 11.4 & 7.6  \\
K-means~\cite{krahenbuhl2015data} & 11.6 & 14.9 & 12.8 & 12.4 \\
Tracking~\cite{wang2015unsupervised} & 14.1 & 18.7 & 16.2 & 15.1 \\
Patch pos.~\cite{doersch2015unsupervised} & 10.0 & 22.4 & - & - \\
Egomotion~\cite{agrawal2015learning} & 9.1 & 11.3 & - & -  \\
\midrule
ImageNet~\cite{krizhevsky2012imagenet} &  29.8 & 34.0 & 37.8 & 37.8 \\
Places~\cite{zhou2014learning} & \textbf{39.4} & \textbf{42.1} & \textbf{46.1} & \textbf{48.8}  \\
\bottomrule
\end{tabular}
\vspace{0.25em}
\caption{SUN397 accuracy for image classification with supervised (bottom), and self-supervised (middle) methods.}
\label{tab:sun_SVM_mAP}
\end{table}

\subsection{Cross-Modal Retrieval}
\label{sec:cross_modal_exp}

As seen in Fig. \ref{fig:overview}, the final layer of the network projects the images on same representation as text as obtained by the LDA model (Section \ref{sec:text_rep}). Therefore, cross-modal retrieval can be directly done by making use of LDA topic probabilities for text and network final predictions for image. We use KL-divergence as a distance metric to short the samples of the target modality, since both the LDA encoding and the CNN output represent probability distributions. 

Note that our comparisons are made with existing methods with reported performance using ImageNet pre-trained AlexNet \cite{krizhevsky2012imagenet} architecture  for image representations and LDA \cite{blei2003latent} or BoW representations for text.

\subsubsection{Wikipedia}

We use the Wikipedia retrieval dataset \cite{rasiwasia2010new}, which consists of 2,866 image-article pairs split into train and test set of 2,173 and 693 pairs respectively 
Further, each image-document pair is labeled with one of ten semantic classes \cite{rasiwasia2010new}.

In Table \ref{table:multi_modal_retrieval_wiki} we compare our results with supervised and unsupervised multi-modal retrieval methods discussed in ~\cite{wang2016comprehensive} and ~\cite{kang2015cross}. Supervised methods make use of class or categorical information associated with each image-document pair, whereas unsupervised methods do not. All of these methods use LDA for text representation and CNN features from pre-trained CaffeNet, which is trained on ImageNet dataset in a supervised setting. We observe that the self-supervised baseline method outperforms unsupervised approaches, and has competitive performance to supervised methods without using any labeled data.

\begin{table}
\centering
\begin{tabular}{l | c | c | c}
\toprule
Method & \shortstack{Image\\Query} & \shortstack{Text\\Query} & Average \\
\midrule
Ours & $39.10$ & $\textbf{43.40}$ & $\textbf{41.25}$\\
TextTopicNet (Wikipedia) \cite{patel2018texttopicnet}& $37.63$ & $40.25$ & $38.94$\\
TextTopicNet (ImageCLEF) \cite{gomez2017self} & $39.58$ & $38.16$ & $38.87$\\
\midrule
CCA \cite{hardoon2004canonical,rasiwasia2010new}& $19.70$ & $17.84$ & $18.77$ \\
PLS \cite{rosipal2006overview} & $30.55$ & $28.03$ & $29.29$ \\
\midrule
SCM* \cite{rasiwasia2010new}& $37.13$ & $28.23$ & $32.68$ \\
GMMFA* \cite{sharma2012generalized} & $38.74$ & $31.09$ & $34.91$ \\
CCA-3V* \cite{gong2014multi} & $40.49$ & $36.51$ & $38.50$ \\
GMLDA* \cite{sharma2012generalized} & $40.84$ & $36.93$ & $38.88$ \\
LCFS* \cite{wang2013learning}& $41.32$ & $38.45$ & $39.88$ \\
JFSSL* \cite{wang2016joint}& $\textbf{42.79}$ & $39.57$ & $41.18$ \\
\bottomrule
\end{tabular}

\vspace{0.25em}
\caption{Mean average precision (MAP) comparison on Wikipedia dataset \cite{rasiwasia2010new} with supervised (bottom), unsupervised (middle) and self-supervised (top) methods. Methods marked with asterisk make use of document (image-text) class category information.}
\label{table:multi_modal_retrieval_wiki}
\end{table}

In Table \ref{table:multi_modal_retrieval_wiki}, we also observe that our method which leverages global and local contexts for self-supervised training leads to state-of-the-art performance, even when compared to fully supervised approaches. This demonstrates that training a network to predict both the global and local semantic contexts in which it is more probable to appear leads to better learning for retrieval task. Further note that, except ours and TextTopicNet \cite{gomez2017self,patel2018texttopicnet} all the other methods use ImageNet pre-trained network.

\subsubsection{Pascal Sentences} We also evaluate our method on pascal sentences dataset \cite{farhadi2010every} which is a subset of pascal VOC dataset. It contains $1000$ pairs of an image along with several sentences from $20$ categories. While, the other methods randomly split the dataset into $600$ training and $400$ testing samples, we test on all $1000$ samples. This is due to the fact that we do not make use of this dataset for training at any point. 

Table \ref{table:multi_modal_retrieval_pascal} provides an extensive comparison with existing methods. Compared to other retrieval methods that use self-supervised visual representations \cite{gomez2017self,patel2018texttopicnet}, our method achieves $1.6\%$ higher MAP with $\frac{1}{4^{th}}$ the size of training data. This demonstrates the efficacy of jointly using global and local self-supervision signals.

\begin{table}
\centering
\begin{tabular}{l | c | c | c}
\toprule
Method & \shortstack{Image\\Query} & \shortstack{Text\\Query} & Average \\
\midrule
Ours & $32.6$ & $\textbf{36.0}$ & $\textbf{34.3}$ \\
TextTopicNet (Wikipedia)\cite{patel2018texttopicnet}& $30.1$ & $35.2$ & $32.7$\\
TextTopicNet (ImageCLEF) \cite{gomez2017self} & $26.4$ & $31.6$ & $29.0$ \\
\midrule
CCA \cite{hardoon2004canonical,rasiwasia2010new}& $9.90$ & $9.7$ & $9.8$ \\
CFA \cite{li2003multimedia} & $18.7$ & $21.6$ & $20.2$ \\
KCCA (Poly) \cite{hardoon2004canonical} & $20.7$ & $19.1$ & $19.9$ \\
KCCA (RBF) \cite{hardoon2004canonical} & $23.3$ & $24.9$ & $24.1$ \\
Bimodal AE \cite{ngiam2011multimodal} & $24.5$ & $25.6$ & $25.1$ \\
Multimodal DBN \cite{srivastava2012multimodal} & $19.7$ & $18.3$ & $19.0$ \\
Corr-AE \cite{feng2014cross} & $26.8$ & $27.3$ & $27.1$ \\
JRL \cite{zhai2014learning} & $30.0$ & $28.6$ & $29.3$ \\
CMDN \cite{peng2016cross} & $\textbf{33.4}$ & $33.3$ & $33.4$ \\
\bottomrule
\end{tabular}
\vspace{0.25em}
\caption{Mean average precision (MAP) comparison on pascal sentences dataset \cite{farhadi2010every} with supervised image representations (bottom) and self-supervised image representations (top) methods.}
\label{table:multi_modal_retrieval_pascal}
\end{table}

\subsection{Qualitative Retrieval Results}
\label{sec:qualitative_results}
Finally, in this section we provide additional qualitative experiments for an image retrieval task.

\begin{figure*}
    \centering
    \includegraphics[width=\textwidth]{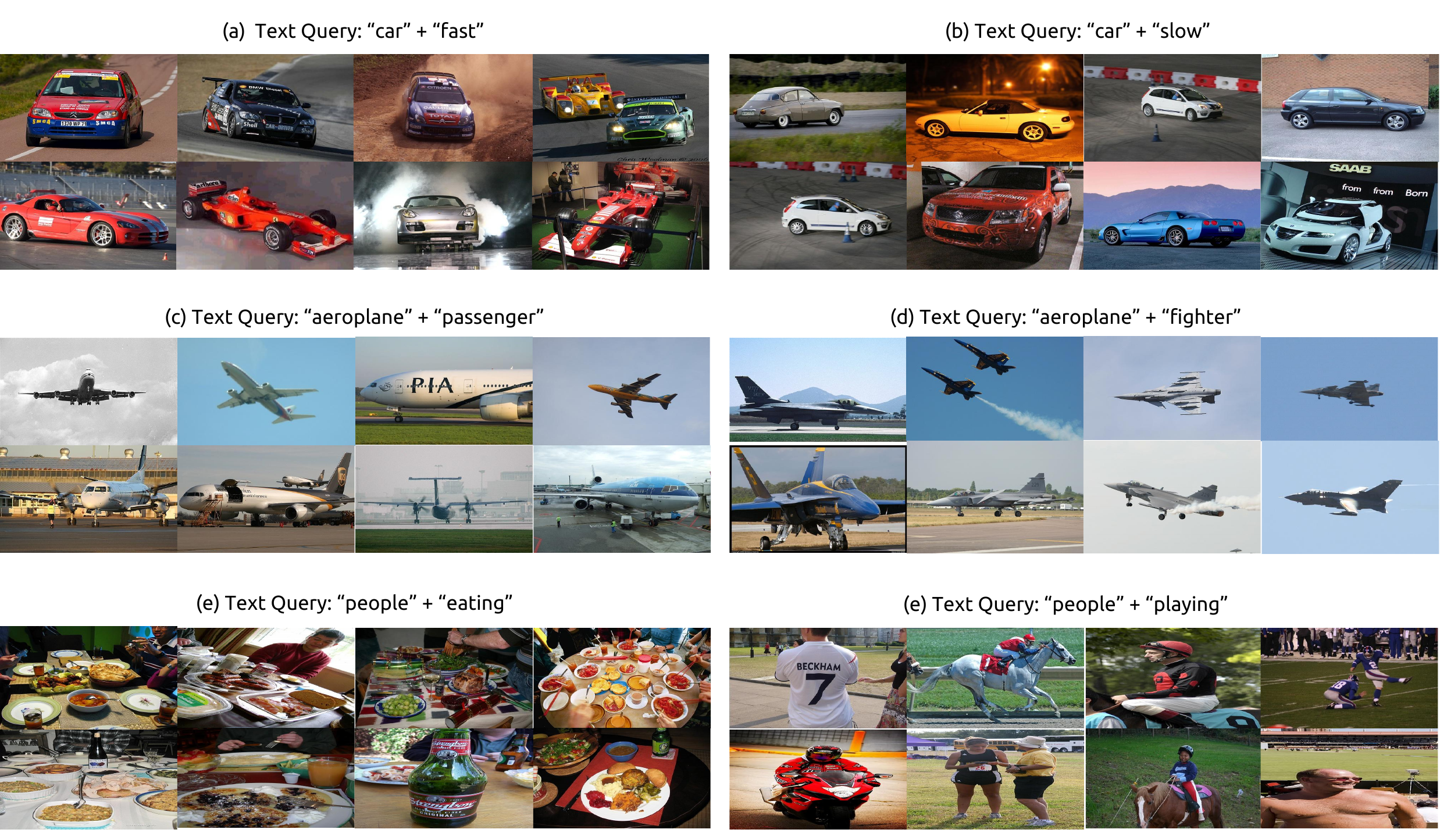}
    \caption{Qualitative examples of text query to image retrieval using nearest neighbour search by comparing network output from caption branch $(f_{C}(x,\Theta))$ with LDA topic probabilities ($\Phi(x^{C})$).}
    \label{fig:text_to_img_qualitative}
\end{figure*}

Figure~\ref{fig:text_to_img_qualitative} shows the top-8 nearest neighbors for a given text query (from left to right and top to bottom: ``car''+``fast'', ``car''+``slow'', ``aeroplane''+``passenger'', ``aeroplane''+``fighter'', ``people''+``eating'', and ``people''+``playing'') in the learned topic space of of our model (without fine tuning). We appreciate that, by leveraging textual semantic information, our method learns rich visual representations that can disambiguate correctly between those combined queries. 

\begin{figure*}
    \centering
    \includegraphics[width=\textwidth]{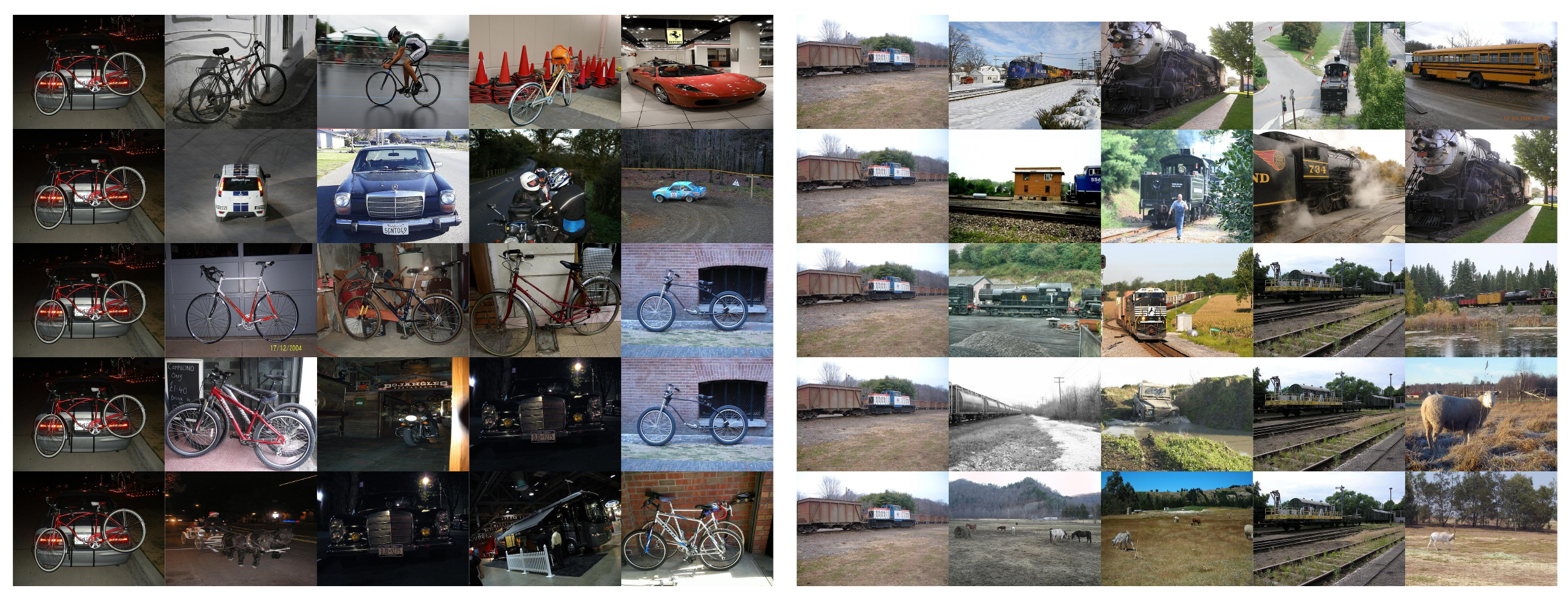}
    \caption{Top 4 nearest neighbors for a given query image image (left-most). Each row makes use of features obtained from different layers of our network (without fine tuning). From top to bottom: \textit{prob-article} $(f_{A}(x,\Theta))$, \textit{prob-caption} $(f_{C}(x,\Theta))$, \textit{fc7}, \textit{fc6}, \textit{pool5}.}
    \label{fig:img_to_img_qualitative}
\end{figure*}

Figure~\ref{fig:img_to_img_qualitative} shows the 4 nearest neighbors for a given query image (left-most), where each row makes use of features obtained from different layers of our model (again without fine tuning).  Query images are randomly selected from PASCAL VOC 2007 dataset and never shown at training time. It can be appreciated that when retrieval is performed in the semantic space layers (prob-article and prob-caption), the results are semantically close, although not necessarily visually similar. As features from earlier layers are used, the results tend to be more visually similar to the query image.

\section{Conclusion}
\label{sec:conclusion}

In this article we put forward a self-supervised method that takes advantage of the natural correlation between an article's text and the images used to illustrate it, in order to learn useful visual representations.

The proposed method is capable of exploiting the rich semantics and broad coverage of illustrated articles, making use of both article-wide semantics and specific image semantics captured by the image caption.

We demonstrated that the learned visual features can transfer well to any general computer vision task such as image classification or object detection, while they can be directly used in a cross-modal retrieval framework yielding state of the art results both on the Wikipedia retrieval dataset and the Pascal Sentences dataset. Notably, the obtained model improves the state of the art not only in comparison to other self-supervised methods, but also when compared to supervised models.

\clearpage
\bibliographystyle{ACM-Reference-Format}
\bibliography{sample-bibliography}

\end{document}